\documentclass[12pt]{extarticle}

\usepackage[english]{babel}
\usepackage{CJKutf8}

\usepackage[letterpaper,top=2cm,bottom=2cm,left=3cm,right=3cm,marginparwidth=1.75cm]{geometry}

\usepackage{amsmath}
\usepackage{graphicx}
\usepackage[colorlinks=true, allcolors=blue]{hyperref}
\usepackage{array}
\usepackage{booktabs}
\usepackage{tabularx}
\usepackage[utf8]{inputenc}
\usepackage{tabularx}
\usepackage{geometry}
\usepackage{float}
\usepackage{multirow}
\usepackage{subcaption}
\usepackage{caption}
\usepackage{longtable}
\usepackage{ragged2e}
\usepackage{colortbl}

\title{\fontsize{18pt}{12pt}\selectfont \textbf{Large Language Models for Semantic Monitoring of Corporate Disclosures: A Case Study on Korea's Top 50 KOSPI Companies}}

\author{
    Junwon Sung\thanks{jwsung@egap.co.kr}\textsuperscript{*},
    Woojin Heo\thanks{wjheo@egap.co.kr},
    Yunkyung Byun\thanks{ykbyun@egap.co.kr},
    Youngsam Kim\thanks{yskim@egap.co.kr}
    \\ 
    \textit{EG Asset Pricing, Republic of Korea} 
}

\date{}

\begin{document}
\maketitle

\begin{abstract}
In the rapidly advancing domain of artificial intelligence, state-of-the-art language models such as OpenAI's GPT-3.5-turbo and GPT-4 offer unprecedented opportunities for automating complex tasks. This research paper delves into the capabilities of these models for semantically analyzing corporate disclosures in the Korean context, specifically for timely disclosure. The study focuses on the top 50 publicly traded companies listed on the Korean KOSPI, based on market capitalization, and scrutinizes their monthly disclosure summaries over a period of 17 months. Each summary was assigned a sentiment rating on a scale ranging from 1(very negative) to 5(very positive). To gauge the effectiveness of the language models, their sentiment ratings were compared with those generated by human experts. Our findings reveal a notable performance disparity between GPT-3.5-turbo and GPT-4, with the latter demonstrating significant accuracy in human evaluation tests. The Spearman correlation coefficient was registered at 0.61, while the simple concordance rate was recorded at 0.82. This research contributes valuable insights into the evaluative characteristics of GPT models, thereby laying the groundwork for future innovations in the field of automated semantic monitoring.
\end{abstract}

\section{Introduction}
In recent years, large language models like OpenAI's ChatGPT have caught people's attention \cite{Heaven2023mit}. What sets these models apart is their ability to learn from the context they're given, making them incredibly versatile when it comes to handling new data \cite{brown2020language, openai2023gpt4}. This means they are great for a wide range of language tasks, without needing specialized training data or fine-tuning \cite{kim2023gpt4, kim2023bloated, lopez2023gpt35, zhang2023large}. 

This is especially useful for tasks that require keeping up with constantly changing information, like monitoring the news in real-time. In our study, we are using these advanced models to analyze the sentiment in corporate announcements. This is an intricate task because these announcements are usually filled with complex data that takes time and expertise to understand. We are specifically looking at how well OpenAI's latest models, like GPT-3.5-turbo and GPT-4, can perform sentiment analysis on corporate announcements from Korean companies. 

Conceptually, corporate disclosure can be broadly categorized into two types: periodic and continuous disclosures \cite{Yang2013}. Periodic disclosures encompass routine reports that companies are obligated to submit to the pertinent regulatory bodies. These reports often encompass financial statements, quarterly updates, and annual summaries. Conversely, continuous disclosures encompass significant information that emerges outside the regular reporting timetable and could potentially influence the valuation of a company's shares or other financial instruments. In South Korea, these ongoing reports are termed `timely disclosures,' while in the United States, they are denoted as `current disclosures' or simply 8-K reports. 

Our monitoring system focuses on continuous disclosures, given their often immediate impact on a company's financial standing and, subsequently, investor decisions. This type of disclosure may include, but is not limited to, merger announcements, changes in executive leadership, regulatory investigations, or any other material events that could affect the stock price or investor perception. Given the increasing importance of continuous corporate disclosure in shaping investor sentiment and market dynamics, the ability to rapidly and accurately assess these communications is crucial. Traditional methods of sentiment analysis often involve manual annotations or rely on simpler algorithms that may not capture the complex events often found in these announcements \cite{ravula2021bankruptcy, bapat2022sentiment}. The limitations of these approaches become even more pronounced when dealing with non-English languages, where cultural and linguistic subtleties can significantly impact the interpretation of sentiment.

Our study aims to fill a gap in the existing literature by leveraging the capabilities of state-of-the-art language models to perform sentiment analysis on Korean corporate announcements. We seek to answer the following research questions:

\begin{enumerate}
    \item How effective are large language models, specifically GPT-3.5-turbo and GPT-4, in analyzing sentiment in corporate announcements?
    \item What challenges or limitations are associated with using large language models like GPT-3.5-turbo and GPT-4 for sentiment analysis in the context of corporate announcements?
\end{enumerate}

The second question aims to uncover any potential drawbacks or limitations of using these advanced models for this specific task. While these models offer a range of capabilities, they are not without challenges. These could include computational constraints or the risk of model biases affecting the analysis. Understanding these challenges is essential for evaluating the practicality and reliability of using large language models for real-time semantic monitoring.

By addressing these research questions, we aim to offer a balanced view of both the capabilities and limitations of using large language models for sentiment analysis in Korean corporate announcements. This should contribute to both academic discussions and provide actionable insights for industry practitioners.

\section{Methodology}

\subsection{Data Collection}
\begin{itemize}
    \item \textbf{Targeted Enterprises:} The top 50 companies by market capitalization were selected from the Korea KOSPI as of June 28, 2023. This decision was based on their significant influence on the market and the general interest in these leading companies.
    \item \textbf{Data Collection Period:} From January 1, 2022, to May 31, 2023.
    \item \textbf{Data Collection Methodology:} The dataset for this study was sourced from the Korea Investor's Network for Disclosure System (KIND). To maintain coherence and relevance, the disclosures were summarized into single sentences using the GPT-3.5 model prior to the sentiment rating process. As noted in the introduction, periodic disclosures were excluded. Specifically, fair reports, business reports, semi-annual reports, and quarterly reports were not included. It is worth noting that the average token length for these periodic disclosures is 65,657, whereas the token length for timely disclosures averages 2,172.
\end{itemize}

\subsection{Data Preprocessing}
Data obtained from corporate disclosures was systematically converted into a monthly time-series format to enable effective sentiment analysis, as depicted in Table \ref{tab:table2}. Each data point includes key elements such as the date, time, title, and a succinct summary of the disclosure's content. The collection process is limited to a maximum of 15 disclosures per month for each company. In cases where a company releases more than 15 disclosures within a single month, only the most recent 15 are chosen for analysis. This limitation is imposed primarily due to the context-length constraints of the language models being used. For example, if a company were to issue daily disclosures from June 1 to June 30, analytically, it would be more valuable to consider the disclosures from June 16 to June 30 rather than the initial 15 from June 1 to June 15.

\subsection{Utilization of the GPT Model}
The GPT model was directed to assign scores between 1 and 5 based on criteria delineated in Table \ref{tab:table1}.

\begin{table}[H]
\centering
\begin{tabularx}{\textwidth}{|c|X|}
\hline
\textbf{Score} & \textbf{Criteria} \\
\hline
1 (Very Negative) & The company's overall situation is very unfavorable, indicating a decline in revenue and profit. Financial conditions are unstable, market share is decreasing, and there are concerns about the ability of management and social responsibility. The future outlook in this situation is highly uncertain, facing threats to the company's sustainability. \\
\hline
2 (Negative) & The company's condition is unfavorable, but certain improvements are possible. The trend of declining revenue and profit continues, and financial conditions are unstable. Market share may vary depending on competitive situations, and evidence of innovation or growth potential is limited. The outlook for the future is not very bright. \\
\hline
3 (Neutral) & The company's situation has not changed significantly, indicating that revenue and profit are stable. Financial conditions are stable, and competitiveness in the market is consistently maintained. Innovation and growth potential are average, and the future outlook remains stable without significant changes. \\
\hline
4 (Positive) & The company is showing significant revenue and growth, indicating that it is being operated well overall. Financial conditions are positive, and there is a trend of increasing market share. There are positive expectations regarding innovation and growth potential, and the outlook for the future is positive. \\
\hline
5 (Very Positive) & The company is achieving explosive revenue and profit, occupying an outstanding position in the market as a result. Financial conditions are very stable, and market share is dominant. The company possesses excellent innovation and growth potential, and the expectations for the future are very high. \\
\hline
\end{tabularx}
\caption{Scoring criteria (common for GPT model and human participants)}
\label{tab:table1}
\end{table}

As illustrated in Table \ref{tab:table1}, the model was prompted to evaluate several key factors, including the company's financial health, market share, and growth potential. Using the OpenAI API, both the prompt (Table \ref{tab:table1}) and summarized data (Table \ref{tab:table2}) were dispatched to the model, from which the GPT's rating was then retrieved.

\begin{table} [H]
    \centering
    \begin{tabularx}{\textwidth}{|c|c|X|}
    \hline
    Date & Time & Details \\
    \hline
    2023-06-13 & 16:30 & Additional Listing (Domestic CB Conversion): CJ CGV Co., Ltd. has additionally listed 383 registered common shares. The issuance price for the 6th time is 26,600 KRW, and for the 9th time is 22,000 KRW. The issuance period is from May 16 to 31, 2023. The dividend base date is January 1, 2023. The capital increase method is a domestic CB conversion, and the listing date is June 16, 2023. \\
    \hline
    2023-06-20 & 15:49 & Capital Increase Decision: CJ CGV decided through a board meeting on June 20, 2023, to increase capital by issuing a total of 74,700,000 common shares. They plan to raise 100 billion KRW for facility funds, 90 billion KRW for operational funds, and 380 billion KRW for debt repayment. The issue price of new shares is 7,630 KRW per share. The assignment date for the new shares is July 31, 2023, and the listing date is expected to be September 27, 2023. \\
    \hline
    2023-06-20 & 16:04 & Loan Decision: A decision was made to lend money to CJ CGV's Hong Kong corporation, CGI HOLDINGS LIMITED. The loan amount is 102.456 billion KRW with an interest rate of 7.37\%. The loan period is from June 20, 2023, to December 20, 2023. This loan is an extension of an existing loan for the improvement of the subsidiary's financial structure. The board decision date is June 20, 2023. \\
    \hline
    2023-06-20 & 16:09 & Transactions with Affiliates: CJ CGV conducted product and service transactions with its affiliate, CJ OliveNetworks, in the third quarter of 2023. The transaction amount totals 12.349 billion KRW, which is 1.75\% of the previous fiscal year's sales. The transaction details include software and other service contracts, and the contract method is by mutual agreement. \\
    \hline
    2023-06-20 & 16:12 & Bond Warning: CJ CGV 35CB (new type) (KR6079161C75) is designated as a bond of concern for investment. This is due to its closing price falling below 80\% of its face value. \\
    \hline
    2023-06-27 & 17:04 & Additional Listing (Domestic CB Conversion): CJ CGV has additionally listed 378 registered common shares. The issuance prices for the 6th and 9th issues are 26,600 KRW and 22,000 KRW, respectively. The issuance date ranges from June 8 to 15, 2023. The dividend calculation date is January 1, 2023. The method of capital increase is domestic CB conversion, and the listing date is June 30, 2023. \\
    \hline
    2023-06-30 & 15:50 & Decision to Provide Collateral for Others: The company decided to provide collateral for CGI Holdings Limited's borrowings of 26.256 billion KRW from KEB Hana Bank Hongkong Branch. The collateral amount is 29.343 billion KRW, which is 7.46\% of the company's equity capital of 393.089 billion KRW. The collateral provision period is from June 30 to September 27, 2023, and the collateral property is in KRW deposit. \\
    \hline
    \end{tabularx}
    \caption{CJ CGV prompt data as of end of June}
    \label{tab:table2}
\end{table}

To ensure a uniform and comprehensive analysis, responses from the GPT models are formatted as depicted in Table \ref{tab:table3}. For enhanced interpretability, each generated response is structured to incorporate a brief rationale along with its corresponding evaluation score.

\begin{table} [H]
    \centering
    \begin{tabularx}{\textwidth}{|c|X|}
        \hline
        \textbf{Rating Score} & \textbf{Reasons for the score} \\
        \hline
        2 (Negative) & CJ CGV is making efforts to secure funds through additional listing and paid-in capital increases. However, given the extension of loans to affiliates, the use of capital increase funds for debt repayment, and the forecast for designation as an investment cautionary stock, the company's financial status is perceived as unstable. Such circumstances could increase uncertainty about the company's future growth and potentially weaken its competitiveness. \\
        \hline
    \end{tabularx}
    \caption{Response example from the GPT-4 model.}
    \label{tab:table3}
\end{table}

\subsection{Evaluation Method}
In our research, we specifically selected the two language models:

\begin{itemize}
    \item \textbf{ChatGPT-3.5-turbo-16K}
    \item \textbf{GPT-4}
\end{itemize}

For the evaluation, both the GPT models and human assessors scrutinized a set of 815 evaluation queries related to the top 50 companies listed on the KOSPI index. These queries were assessed using a standardized 1-5 point scale. To ensure consistency, the prompted queries were kept uniform across all evaluations. It should be noted that the queries were translated into English to optimize the performance of the model, a technique commonly employed to enhance accuracy.

\begin{figure}[H]
\centering
\includegraphics[width=1.0\linewidth]{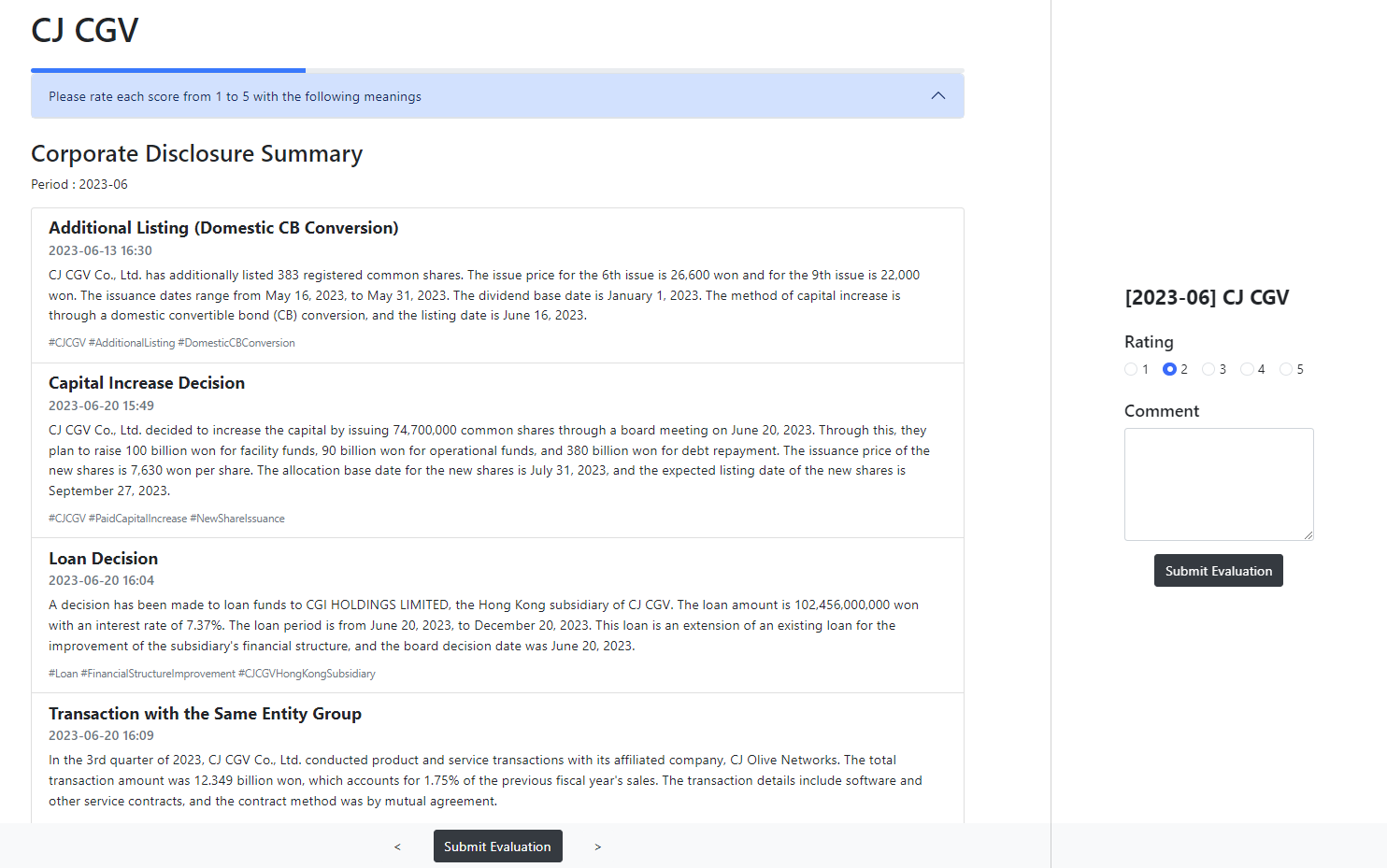} 
\caption{Disclosure rating screen for human raters}
\label{fig:fig1}
\end{figure}

For human evaluation, we collaborated with two highly proficient experts boasting over a decade of experience within the realm of financial data analysis. These experts conducted individual assessments of the data, employing identical criteria to those utilized for evaluating the GPT models. To ensure a fluid and efficient evaluation process, we equipped the experts with a user-friendly web-based interface, as depicted in Figure \ref{fig:fig1}.

We utilized Cohen's Kappa statistic to measure the degree of consistency between human evaluators, yielding a value of 0.352. This result indicates a fair level of inter-rater agreement. Additionally, the simple agreement rate was observed to be 68\%, further substantiating the reliability of the assessment. The scores attributed to each query by the two experts were summed and subsequently averaged. Any fractional values resulting from this averaging process were methodically rounded down to the nearest tenth, reflecting a conservative approach to the rating values.

\section{Experimental Results}

\subsection{Conditions for Rating Adjustments}
Previous study has shown that GPT model summaries of public disclosures could emphasize a positive or negative tone more than human intuition might suggest \cite{kim2023bloated}. Therefore, this study considered the possibility of bias in the sentiment score assigned by the GPT model. To evaluate the effect of this potential bias, we constructed several artificial rating adjustment conditions as follows:

\begin{itemize}
    \item \textbf{Condition 1:} No adjustments. This serves as a control group, with no adjustments made, to have a baseline performance metric.
    \item \textbf{Condition 2:} Subtract 1 point if the GPT score is 4 or above. This adjustment helps balance the model's tendency to be overly positive, making its evaluations more aligned with human judgment.
    \item \textbf{Condition 3:} Add 1 point if the GPT score is 2 or lower. By adding a point for lower scores, you are testing the reverse hypothesis – that the model might be unduly negative or understate negative tones.
    \item \textbf{Condition 4:} Add 1 point if the GPT score is 2 or below, and subtract 1 point if the score is 4 or above. This combines Conditions 2 and 3 to examine whether the model has both overestimation and underestimation biases at the same time.
\end{itemize}

\subsection{Results of the Conditions}
The correlation and concordance rate results for each condition are presented in Table \ref{tab:table4}. In Condition 1, The concordance rate is higher in ChatGPT-3.5 than in GPT-4, yet GPT-4 has a higher Spearman and Kendall coefficient. The GPT-4 model records the highest correlation in Condition 2 (0.82 agreement rate, 0.61 Spearman, and 0.59 Kendall). However, both GPT models show reduced correlations compared to the baseline condition in Condition 3 and Condition 4.

The key findings from these results are summarized as follows:

\begin{itemize}
    \item \textbf{The GPT-4 model demonstrates higher correlations than GPT-3.5 in all conditions.}
    \item \textbf{The GPT-4 model in Condition 2 shows the highest performance across all settings for every measure.}
    \item \textbf{The effects of rating adjustments are clearly observed in correlation measures.}
\end{itemize}

\begin{figure}[H] 
    \centering
    \includegraphics[width=0.72\textwidth]{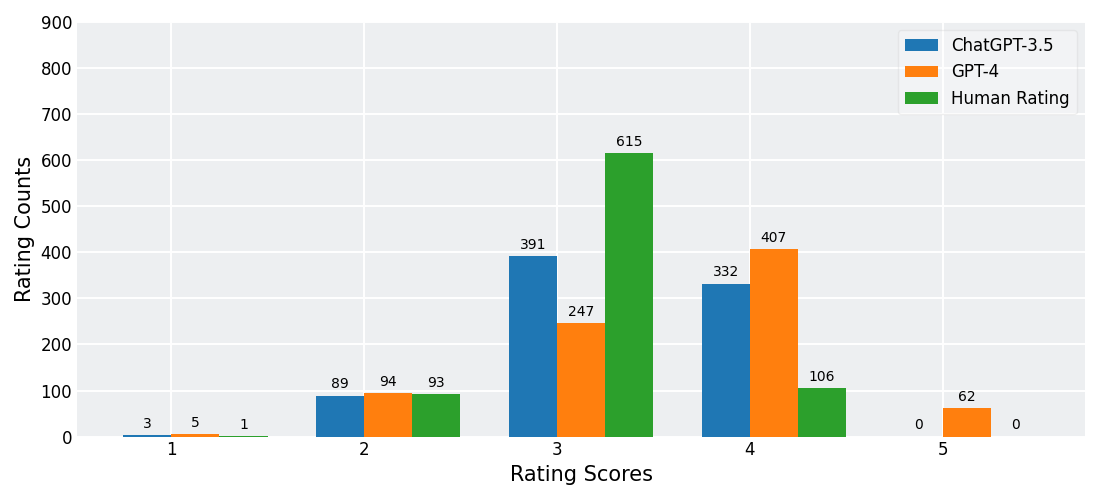}
    \caption{Condition 1}
    \label{fig:fig2}
    \vspace{0.25cm}
    \includegraphics[width=0.72\textwidth]{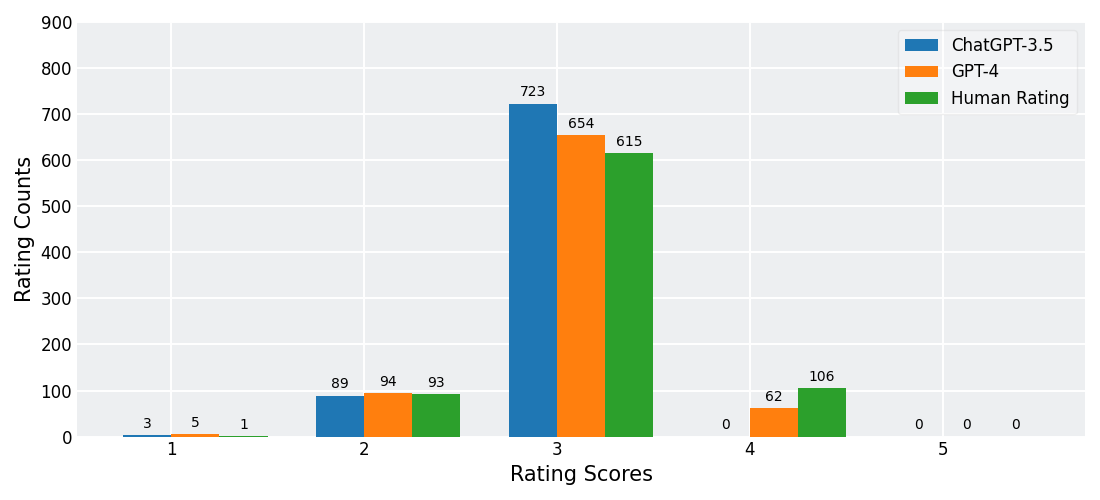}
    \caption{Condition 2}
    \label{fig:fig3}
    \vspace{0.25cm}
    \includegraphics[width=0.72\textwidth]{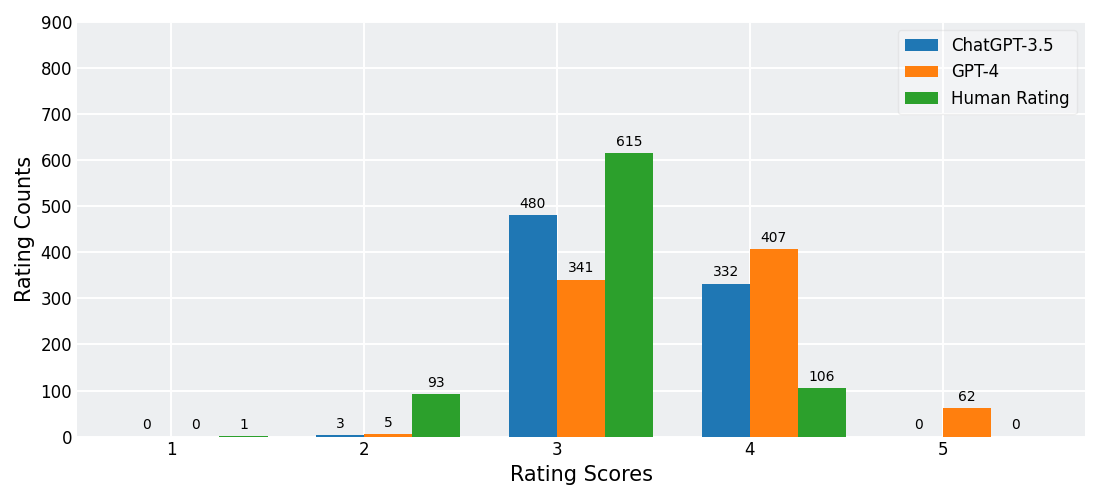}
    \caption{Condition 3}
    \label{fig:fig4}
    \vspace{0.25cm}
    \includegraphics[width=0.72\textwidth]{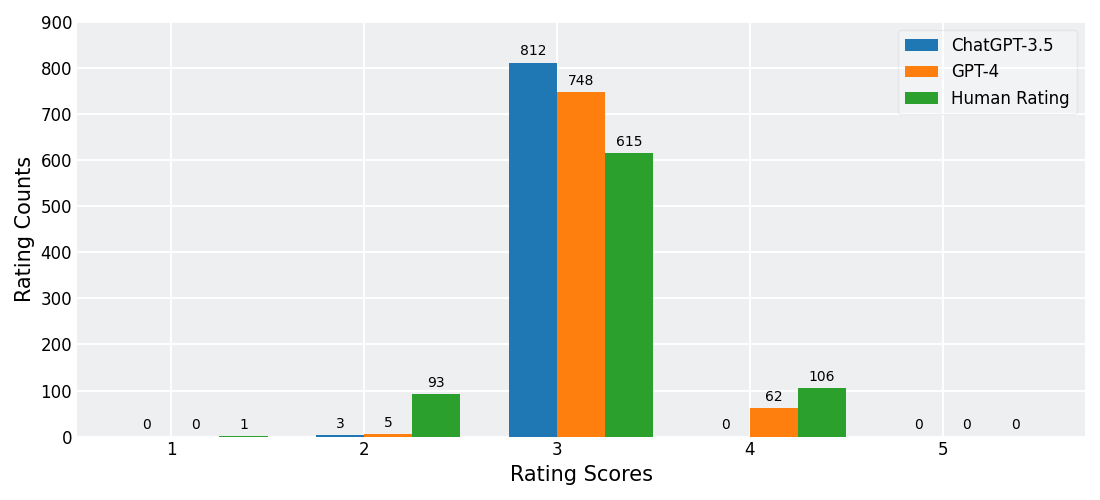}
    \caption{Condition 4}
    \label{fig:fig5}
\end{figure}

The third point on the previous list emphasizes a significant imbalance within the rating scales. As depicted in Figure \ref{fig:fig2}, human evaluators assigned roughly 75\%(615/815) of their ratings to a score of 3. This skew allowed the GPT models to achieve relatively high performance in terms of concordance rate under Condition 4, even while registering the lowest correlation values.

The performance pattern of the ChatGPT-3.5 model differs subtly from that of the GPT-4 model. Specifically, ChatGPT-3.5 achieves its peak concordance rate under Condition 2, while it registers the highest correlation values in Condition 1. This variation could suggest that the ChatGPT-3.5 model's ratings are less consistent than those generated by the GPT-4 model. For a detailed view of the rating distribution across various conditions for the models, please refer to Figures \ref{fig:fig2} through \ref{fig:fig5}.

\begin{table}[H]
\centering
\begin{tabularx}{\textwidth}{>{\RaggedRight}Xcccc}
\toprule
\textbf{Condition} & \textbf{GPT Model} & \textbf{Concordance Rate} & \textbf{Spearman} & \textbf{Kendall} \\
\midrule
Condition 1 & ChatGPT-3.5 & 0.60 & 0.48 & 0.46 \\
& GPT-4 & 0.43 & 0.57 & 0.54 \\
\midrule
Condition 2 & ChatGPT-3.5 & 0.77 & 0.42 & 0.41 \\
& \textbf{GPT-4} & \textbf{0.82} & \textbf{0.61} & \textbf{0.59} \\
\midrule
Condition 3 & ChatGPT-3.5 & 0.59 & 0.40 & 0.39 \\
& GPT-4 & 0.39 & 0.51 & 0.48 \\
\midrule
Condition 4 & ChatGPT-3.5 & 0.76 & 0.13 & 0.13 \\
& GPT-4 & 0.79 & 0.42 & 0.40 \\
\bottomrule
\end{tabularx}
\caption{Performance of GPT models}
\label{tab:table4}
\end{table}

\section{Discussions}

\textbf{ChatGPT-3.5 VS. GPT-4.} While ChatGPT-3.5 shows a higher concordance rate than the GPT-4 model in Condition 1 (0.6 versus 0.43), GPT-4 ultimately outshines ChatGPT-3.5 when comparing their peak performance rates in Condition 2 (0.82 versus 0.77). Notably, the correlations between GPT-4 and human ratings consistently outperform those of ChatGPT-3.5, indicating that GPT-4 offers superior consistency. This outcome is not unexpected, given GPT-4's superior performance records in tasks related to common sense reasoning \cite{openai2023gpt4}.
\\
\\
\textbf{Rating Adjustments.} The empirical finding \cite{kim2023bloated} that the GPT models might over-emphasize positive tones seems validated in our results. Figures 2 and 3 clearly illustrate that both GPT models have a tendency to overestimate positive disclosures when compared to human evaluators. Interestingly, this tendency is not symmetrical. As a result, Conditions 4 and 5 yielded suboptimal performance for the GPT models overall. The GPT-4 model under Condition 2 aligns most closely with the human evaluators and exhibits the highest levels of accuracy and correlations in our results. Our empirical study suggests that prompt engineering alone may be insufficient to address the domain-specific nuances of financial texts like corporate disclosures. This is because large language models may struggle to accurately estimate the impact of such financial events.
\\
\\
\textbf{Limitations and Challenges.} While this study marks a significant stride in the application of Large Language Models (LLMs) like GPT to corporate disclosures, several challenges and limitations must be acknowledged. First and foremost, the GPT models operated without the advantage of background knowledge about the companies being analyzed. This presented a significant drawback, as it limited the depth of their analysis and their ability to understand context. Without knowledge of a company's history, market position, or unique financial complexities, the insights generated by the models were inherently limited. 

Secondly, the absence of external sources such as financial data in tables or news articles further constrained the models' performance. Financial reports are often intricate documents accompanied by various supplementary data and market analyses. The inability to integrate and interpret these additional resources may have prevented a more comprehensive and nuanced evaluation of corporate disclosures. Furthermore, the mathematical capabilities of the GPT models present another set of limitations. While they can perform basic arithmetic and some algebraic operations, their ability to understand and analyze complex financial formulas or perform advanced statistical analyses is limited \cite{frieder2023mathematical}. This is especially pertinent when evaluating corporate financial reports, which often require a sophisticated understanding of accounting principles and mathematical models for accurate interpretation. However, it is worth noting that some of these limitations can be mitigated by employing expert libraries for the problems that GPT models struggle with \cite{yao2023react}. This hybrid approach could potentially offer a more accurate and nuanced analysis, bringing together the text-processing strengths of GPT models with the computational rigor of specialized libraries.

Lastly, LLMs like ChatGPT are subject to modifications, which could lead to inconsistent performance \cite{chen2023chatgpts}. This inherent risk of variability must be addressed to ensure the stability of any monitoring system that relies on these models.
\\
\\
\textbf{Future Work.} There is significant room for growth and expansion of this research. Potential avenues include:
\begin{itemize}
    \item \textbf{Incorporating the GPT-4 model for the summarization process, hypothesizing that its advanced capabilities might yield even more accurate summaries than ChatGPT-3.5-turbo.}
    \item \textbf{Developing mechanisms to feed contextual information about companies to the GPT models, allowing for a richer and more nuanced analysis.}
    \item \textbf{Incorporating external data sources into the analysis, such as financial data tables and news articles, for a more comprehensive sentiment analysis.}
\end{itemize}

\section{Conclusion}
In conclusion, our study demonstrates the potential of large language models, such as ChatGPT-3.5-turbo and GPT-4, in performing sentiment analysis on corporate announcements from Korea's top 50 KOSPI companies. These models exhibit varying degrees of success in evaluating the sentiment of continuous disclosures. Through empirical analysis, we observed that GPT-4 outperforms ChatGPT-3.5-turbo in terms of both consistency and correlation with human evaluations, indicating its superior ability to comprehend and assess complex financial language. The study also highlights the importance of addressing biases and limitations in LLMs' output, as evidenced by the need for rating adjustments to align model-generated scores with human judgment. However, challenges like the lack of background knowledge, the absence of external data integration remind us of the evolving nature of LLMs' capabilities. Future research should explore approaches to improve contextual understanding, integrate external data, and enhance mathematical analysis within the framework of sentiment monitoring in the financial domain. This research contributes to a deeper understanding of the capabilities and limitations of LLMs for real-time semantic monitoring and presents valuable insights for both academia and industry practitioners engaged in sentiment analysis of corporate disclosures.

\bibliography{references}

\newpage
\appendix
\section*{\fontsize{15pt}{15pt}\selectfont Appendix: Individual results of the KOSPI 50 Companies}

* Concordance Rate and correlation index of the companies in condition 2.
\\
* NaN: When assessing the rating relationship between humans and models, identical values on one or both sides prevent accurate ranking and can lead to undefined correlation coefficients like Spearman or Kendall.

\begin{longtable}{|l|c|c|c|}
\hline
\textbf{Company Name} & \textbf{Concordance Rate} & \textbf{Spearman} & \textbf{Kendall} \\
\hline
HD Korea Shipbuilding & 0.65 & 0.31 & 0.3 \\
HD Hyundai Heavy Industries & 0.71 & 0.56 & 0.55 \\
HMM & 0.92 & 0.85 & 0.83 \\
KB Financial & 0.82 & 0.64 & 0.62 \\
LG & 0.75 & -0.04 & -0.04 \\
LG Household \& Health Care & 0.81 & 0.74 & 0.72 \\
LG Energy Solution & 0.69 & NaN & NaN \\
LG Innotek & 0.82 & 0.7 & 0.69 \\
LG Electronics & 0.88 & 0.58 & 0.56 \\
LG Chem & 0.82 & 0.71 & 0.7 \\
NAVER & 0.94 & NaN & NaN \\
POSCO Holdings & 0.82 & 0.77 & 0.74 \\
S-Oil & 0.71 & 0.59 & 0.57 \\
SK & 0.88 & NaN & NaN \\
SK Innovation & 0.69 & 0.39 & 0.38 \\
SK Telecom & 1 & NaN & NaN \\
SK Hynix & 0.94 & 0.91 & 0.9 \\
Korea Zinc & 0.88 & 0.71 & 0.7 \\
Kia & 0.65 & 0.34 & 0.32 \\
IBK & 0.75 & 0.58 & 0.56 \\
Daewoo Shipbuilding & 0.71 & 0.5 & 0.47 \\
Korean Air & 0.88 & 0.77 & 0.76 \\
Doosan Mobility Innovation & 0.56 & 0.28 & 0.28 \\
Lotte Chemical & 0.82 & 0.31 & 0.31 \\
Meritz Financial & 0.81 & 0.45 & 0.45 \\
Samsung SDI & 0.88 & 0.69 & 0.67 \\
Samsung C\&T & 0.76 & -0.04 & -0.04 \\
Samsung Biologics & 0.65 & 0.39 & 0.39 \\
Samsung Life & 0.94 & 0.82 & 0.81 \\
Samsung SDS & 0.88 & 0.69 & 0.68 \\
Samsung Electro-Mechanics & 0.88 & 0.77 & 0.75 \\
Samsung Electronics & 0.94 & 0.89 & 0.87 \\
Samsung Fire \& Marine Insurance & 0.88 & 0.77 & 0.76 \\
Celltrion & 0.76 & 0.74 & 0.72 \\
Shinhan Financial & 0.88 & 0.69 & 0.68 \\
Woori Financial & 0.94 & 0.86 & 0.86 \\
Kakao & 0.71 & 0.38 & 0.37 \\
Kakao Bank & 0.82 & 0.45 & 0.45 \\
KT & 0.82 & 0.55 & 0.55 \\
KT\&G & 0.92 & 0.92 & 0.9 \\
Krafton & 0.94 & 0.88 & 0.87 \\
POSCO International & 0.71 & 0.28 & 0.26 \\
POSCO Futurem & 0.75 & 0.61 & 0.6 \\
Hana Financial & 0.87 & 0.77 & 0.76 \\
HYBE & 0.92 & 0.88 & 0.87 \\
Korea Electric Power & 0.88 & 0.61 & 0.6 \\
Hanwha Solutions & 0.94 & 0.73 & 0.72 \\
Hyundai Glovis & 0.81 & 0.61 & 0.6 \\
Hyundai Mobis & 0.82 & 0.52 & 0.5 \\
Hyundai Motor & 0.88 & 0.69 & 0.68 \\
\hline
\end{longtable}

\end{document}